\documentclass[nohyperref]{article}

\usepackage{microtype}
\usepackage{graphicx}
\usepackage{booktabs} %

\usepackage{hyperref}

\usepackage[accepted]{icml2023}

\usepackage{amsmath}
\usepackage{amssymb}
\usepackage{mathtools}
\usepackage{amsthm}

\usepackage[capitalize,noabbrev]{cleveref}

\theoremstyle{plain}

\theoremstyle{definition}

\theoremstyle{remark}

\usepackage{pifont}%
\usepackage{xcolor}%
\usepackage{mathtools}%
\usepackage{subcaption}
\usepackage{caption}
\usepackage{adjustbox}
\usepackage{arydshln}
\usepackage{tabularx}
\usepackage{makecell}

\mathtoolsset{showonlyrefs=true} %

\newcommand{\cmark}{\ding{51}}%
\newcommand{\xmark}{\ding{55}}%
\newcommand{\crossmark}{{\color{red} \xmark}}%
\newcommand{\tickmark}{{\color{green} \cmark}}%

\newcommand{\sdval}[2]{$#1 \pm #2$}

\newcommand{\hp}[1]{\hphantom{#1}}

\newcommand{\timeschange}[2]{$\mathit{#1\times - #2\times}$}

\icmltitlerunning{Continuous DEQs: Infinite Time Neural ODEs}

\begin{document}

\twocolumn[
\icmltitle{Continuous Deep Equilibrium Models: Training Neural ODEs faster by integrating them to Infinity}

\icmlsetsymbol{equal}{*}

\begin{icmlauthorlist}
\icmlauthor{Avik Pal}{csail,miteecs}
\icmlauthor{Alan Edelman}{csail,mitmath}
\icmlauthor{Christopher Rackauckas}{csail}
\end{icmlauthorlist}

\icmlaffiliation{csail}{CSAIL MIT}
\icmlaffiliation{mitmath}{Department of Mathematics MIT}
\icmlaffiliation{miteecs}{Department of Electrical Engineering and Computer Science MIT}

\icmlcorrespondingauthor{Avik Pal}{avikpal@mit.edu}

\icmlkeywords{Machine Learning, Implicit Neural Networks, Deep Equilibrium Models, Neural ODEs, ICML}

\vskip 0.3in
]

\printAffiliationsAndNotice{\icmlEqualContribution} %

\begin{abstract}
Implicit models separate the definition of a layer from the description of its solution process. While implicit layers allow features such as depth to adapt to new scenarios and inputs automatically, this adaptivity makes its computational expense challenging to predict. In this manuscript, we \textit{increase the ``implicitness" of the DEQ by redefining the method in terms of an infinite time neural ODE}, which paradoxically decreases the training cost over a standard neural ODE by $\mathit{2} - \mathit{4 \times}$. Additionally, we address the question: \textit{is there a way to simultaneously achieve the robustness of implicit layers while allowing the reduced computational expense of an explicit layer?} To solve this, we develop Skip and Skip Reg. DEQ, an implicit-explicit (IMEX) layer that simultaneously trains an explicit prediction followed by an implicit correction. We show that training this explicit predictor is free and even decreases the training time by $\mathit{1.11} - \mathit{3.19 \times}$. Together, this manuscript shows how bridging the dichotomy of implicit and explicit deep learning can combine the advantages of both techniques.
\end{abstract}

\begin{figure*}
    \centering
    \includegraphics[width=0.9\textwidth]{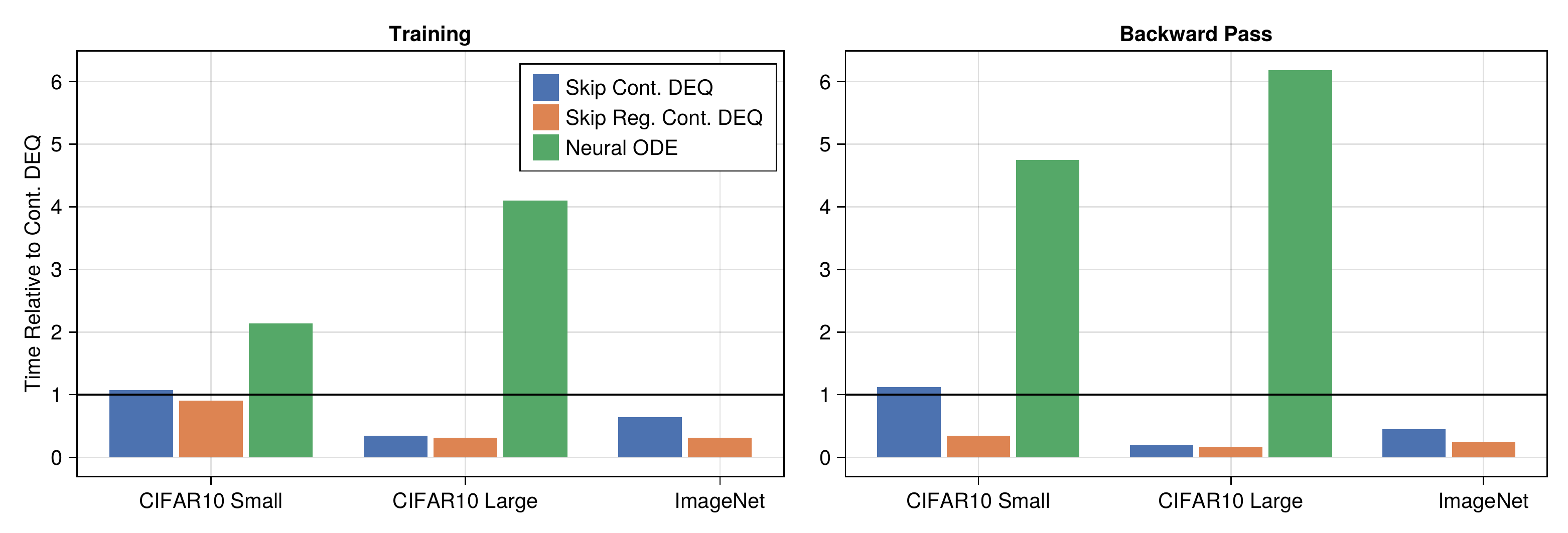}
    \caption{\textbf{Relative Training and Backward Pass Timings against Continuous DEQs} \textit{(lower is better)}: In all scenarios, Neural ODEs take $\mathit{4.7} - \mathit{6.182 \times}$ more time in the backward pass compared to Vanilla Continuous DEQs. Whereas combining Skip (Reg.) with Continuous DEQs accelerates the backward pass by $\mathit{2.8} - \mathit{5.9 \times}$.}
    \label{fig:summary_plot}
\end{figure*}

\section{Introduction}
\label{sec:introduction}

Implicit layer methods, such as Neural ODEs and Deep Equilibrium models \cite{chen2018neural, bai_deep_2019, ghaoui_implicit_2020}, have gained popularity due to their ability to automatically adapt model depth based on the ``complexity'' of new problems and inputs. The forward pass of these methods involves solving steady-state problems, convex optimization problems, differential equations, etc., all defined by neural networks, which can be expensive. However, training these more generalized models has empirically been shown to take significantly more time than traditional explicit models such as recurrent neural networks and transformers. \textit{Nothing within the problem's structure requires expensive training methods, so we asked, can we reformulate continuous implicit models so that this is not the case}?

\citet{grathwohl2018ffjord, dupont2019augmented, kelly2020learning, finlay2020train} have identified several problems with training implicit networks. These models grow in complexity as training progresses, and a single forward pass can take over 100 iterations~\citep{kelly2020learning} even for simple problems like MNIST. Deep Equilibrium Models~\citep{bai_deep_2019, bai_multiscale_2020} have better scaling in the backward pass but are still bottlenecked by slow steady-state convergence. \citet{bai2021stabilizing} quantified several convergence and stability problems with DEQs. They proposed a regularization technique by exploiting the ``implicitness" of DEQs to stabilize their training. \textit{We marry the idea of faster backward pass for DEQs and continuous modeling from Neural ODEs to create Infinite Time Neural ODEs which scale significantly better in the backward pass and drastically reduce the training time}. %

Our main contributions include\footnote{We provide an anonymous version of our codebase \url{https://anonymous.4open.science/r/deq_icml_2023/} with the intent of public release after the review period}
\begin{enumerate}
    \item An improved DEQ architecture (Skip-DEQ) that uses an additional neural network to predict better initial conditions.

    \item A regularization scheme (Skip Regularized DEQ) incentivizes the DEQ to learn simpler dynamics and leads to faster training and prediction. Notably, this does not require nested automatic differentiation and thus is considerably less computationally expensive than other published techniques.

    \item A continuous formulation for DEQs as an infinite time neural ODE, which paradoxically accelerates the backward pass over standard neural ODEs by replacing the continuous adjoints with a simple linear system.

    \item We demonstrate the seamless combination of Continuous DEQs with Skip DEQs to create a drop-in replacement for Neural ODEs without incurring a high training cost.
\end{enumerate}

\section{Background}
\label{sec:background}

\begin{figure}[t]
    \centering
    \includegraphics[width=0.9\linewidth]{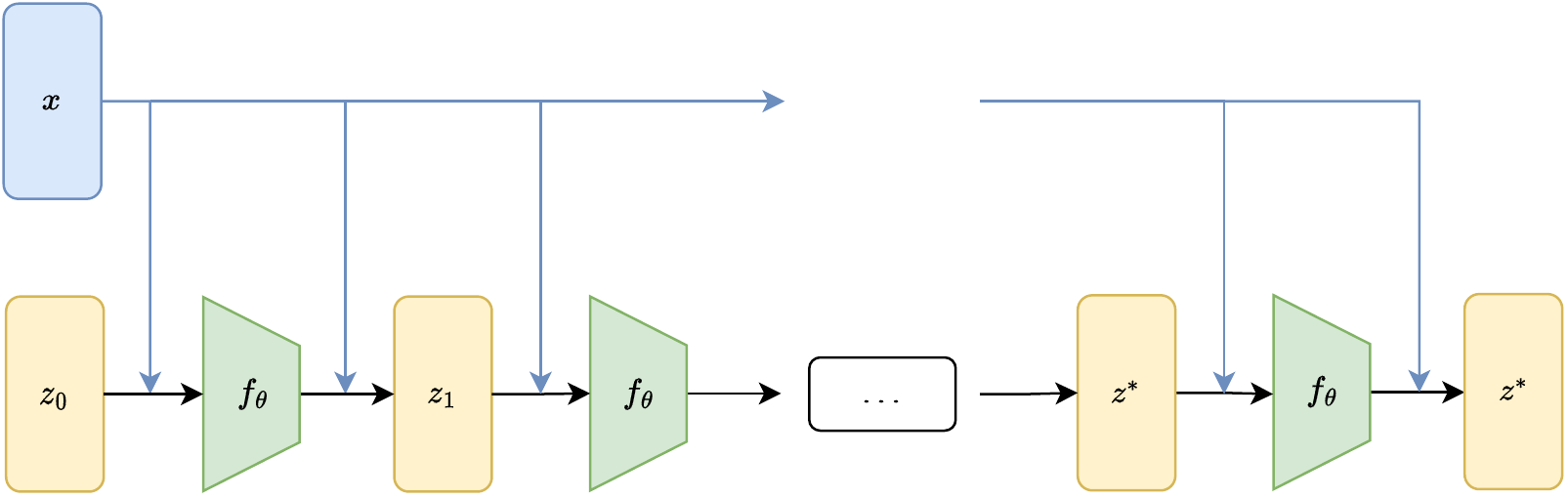}
    \caption{\textbf{Discrete DEQ Formulation}: Discrete DEQ Block where the input $x$ is injected at every iteration till the system (with initial state $z_0$) converges to a steady $z^*$. In Vanilla DEQ, $z_0 = 0$ while in Skip DEQ, an additional explicit model $g_\phi$ (which can potentially share the weights of $f_\theta$) is used to set the initial state $z_0 = g_\phi(x)$.}
    \label{fig:model_architectures}
\end{figure}

Explicit Deep Learning Architectures specify a projection $f: \mathcal{X} \mapsto \mathcal{Z}$ by stacking multiple ``layers". Implicit models, however, define a solution process instead of directly specifying the projection. These methods enforce a constraint on the output space $\mathcal{Z}$ by learning $g: \mathcal{X} \times \mathcal{Z} \mapsto \mathbb{R}^n$. By specifying a solution process, implicit models can effectively vary features like depth to adapt automatically to new scenarios and inputs. Some prominent implicit models include Neural ODEs~\citep{chen2018neural}, where the output $z$ is defined by the ODE $\frac{dz}{dt} = g_\phi(x, t)$. \citet{liu2019neural} generalized this framework to Stochastic Differential Equations (SDEs) by stochastic noise injection, which regularizes the training of Neural ODEs allowing them to be more robust and achieve better generalization. \citet{bai_deep_2019} designed equilibrium models where the output $z$ was constrained to be a steady state, $z^* = f_\theta(z^*, x)$. Another example of implicit layer architectures is seen in \citet{amos2017optnet, agrawal2019differentiable} set $z$ to be the solution of convex optimization problems.

Deep Implicit Models essentially removed the design bottleneck of choosing the ``depth" of neural networks. Instead, these models use a ``tolerance" to determine the accuracy to which the constraint needs to be satisfied. Additionally, many of these models only require $O(1)$ memory for backpropagation, thus alluding to potential increased efficiency over their explicit layer counterparts. However, evaluating these models require solving differential equations~\citep{chen2018neural, liu2019neural}, non-linear equations~\citep{bai_deep_2019}, convex optimization problems~\citep{amos2017optnet, agrawal2019differentiable}, etc. Numerous authors~\citep{dupont2019augmented, grathwohl2018ffjord, finlay2020train, kelly2020learning, ghosh2020steer, bai2021stabilizing} have noted that this solution process makes implicit models significantly slower in practice during training and prediction compared to explicit networks achieving similar accuracy.

\subsection{Neural Ordinary Differential Equations}
\label{subsec:neural_odes}

Initial Value Problems (IVPs) are a class of ODEs that involve finding the state at a later time $t_1$, given the value $z_0$ at time $t_0$. \citet{chen2018neural} proposed the Neural ODE framework, which uses neural networks to model the ODE dynamics
\begin{equation}
    \frac{dz(t)}{dt} = f_{\theta}\left(z\right)
\end{equation}
Using adaptive time stepping allows the model to operate at a variable continuous depth depending on the inputs. Removal of the fixed depth constraint of Residual Networks provides a more expressive framework and offers several advantages in problems like density estimation~\cite{grathwohl2018ffjord}, irregularly spaced time series problems~\cite{rubanova2019latent}, etc. Training Neural ODEs using continuous adjoints come with the added benefit of constant memory overhead. However, this benefit often leads to slower training, since we need to backsolve an ODE. We defer the exact details of the continuous adjoint equations to \citet{chen2018neural}.

\subsection{Deep Equilibrium Models}
\label{subsec:deep_equilibrium_models}

Deep Equilibrium Networks (DEQs)~\citep{bai_deep_2019} are implicit models where the output space represents a steady-state solution. Intuitively, this represents infinitely deep neural networks with input injection, i.e., an infinite composition of explicit layers $z_{n + 1} = f_\theta(z_n, x)$ with $z_0 = 0$ and $n \rightarrow \infty$. In practice, it is equivalent to evaluating a dynamical system until it reaches a steady state $z^* = f_\theta(z^*, x)$. \citet{bai_deep_2019, bai_multiscale_2020} perform nonlinear fixed point iterations of the discrete dynamical system using Broyden's method~\citep{broyden1965class, bai_multiscale_2020} to reach this steady-state solution. 

Evaluating DEQs requires solving a steady-state equation involving multiple evaluations of the explicit layer slowing down the forward pass. However, driving the solution to steady-state makes the backward pass very efficient~\citep{johnson2012notes}. Despite a potentially infinite number of evaluations of $f_\theta$ in the forward pass, backpropagation only requires solving a linear equation.
\begin{gather}
    z^* = f_\theta(z^*, x)\\
    \implies \frac{\partial z^*}{\partial \theta} = \frac{f_\theta(z^*, x)}{\partial z^*} \cdot \frac{\partial z^*}{\partial \theta} + \frac{\partial f_\theta(z^*, x)}{\partial \theta}\\
    \implies \left(I - \frac{\partial f_\theta(z^*, x)}{\partial z^*}\right) \frac{\partial z^*}{\partial \theta} = \frac{\partial f_\theta(z^*, x)}{\partial \theta}
\end{gather}
For backpropagation, we need the Vector-Jacobian Product~(VJP):
\begin{align}
    \left(\frac{\partial z^*}{\partial \theta}\right)^T v &= \left( \frac{\partial f_\theta(z^*, x)}{\partial \theta} \right)^T\left( I - \frac{\partial f_\theta(z^*, x)}{\partial z^*} \right)^{-T} v\\
    &= \left( \frac{\partial f_\theta(z^*, x)}{\partial \theta} \right)^T g
\end{align}
where $v$ is the gradients from layers after the DEQ module. Computing $\left( I - \frac{\partial f_\theta(z^*, x)}{\partial z^*} \right)^{-T}$ is expensive and makes DEQs non-scalable to high-dimensional problems. Instead, we solve the linear equation $ g =  \left(\frac{\partial f_\theta(z^*, x)}{\partial z^*} \right)^{T} g + v$ using Newton-Krylov Methods like GMRES~\citep{saad1986gmres}. To compute the final VJP, we need to compute $\left( \frac{\partial f_\theta(z^*, x)}{\partial \theta} \right)^T g$, which allows us to efficiently perform the backpropagation without explicitly computing the Jacobian.

\subsubsection{Multiscale Deep Equilibrium Network}
\label{subsubsec:mdeq}

Multiscale modeling~\citep{burt1987laplacian} has been the central theme for several deep computer vision applications~\citep{farabet2012learning, yu2015multi, chen2016attention, chen2017deeplab}. The standard DEQ formulation drives a single feature vector to a steady state. \citet{bai_multiscale_2020} proposed Multiscale DEQ (MDEQ) to learn coarse and fine-grained feature representations simultaneously. MDEQs operate at multiple feature scales $z = \left\{ z_1, z_2, \dots, z_n \right\}$, with the new equilibrium state $z^* = f_\theta(z_1^*, z_2^*, \dots, z_n^*, x)$. All the feature vectors in an MDEQ are interdependent and are simultaneously driven to a steady state. \citet{bai_multiscale_2020} used a Limited-Memory Broyden Solver~\citep{broyden1965class} to solve these large scale computer vision problems. We use this MDEQ formulation for all our classification experiments.

\subsubsection{Jacobian Stabilization}
\label{subsubsec:jacobian_stabilization}

Infinite composition of a function $f_\theta$ does not necessarily lead to a steady-state -- chaos, periodicity, divergence, etc., are other possible asymptotic behaviors. The Jacobian Matrix $J_{f_\theta}(z^*)$ controls the existence of a stable steady-state and influences the convergence of DEQs in the forward and backward passes. \citet{bai2021stabilizing} describes how controlling the spectral radius of $J_{f_\theta}(z^*)$ would prevent simpler iterative solvers from diverging or oscillating. \citet{bai2021stabilizing} introduce a Jacobian term to the training objective to regularize the model training. The authors use the Hutchinson estimator~\citep{hutchinson1989stochastic} to compute and regularize the Frobenius norm of the Jacobian.
\begin{equation}
    \mathcal{L}_{jac} = \lambda_{jac} \frac{\| \epsilon^T J_{f_\theta}(z^*) \|_2^2}{d}; \quad \epsilon \sim \mathcal{N}(0, I_d)
\end{equation}
While well-motivated, the disadvantage of this method is that the Hutchinson trace estimator requires automatic differentiation in the loss function, thus requiring higher order differentiation in the training process and greatly increasing the training costs. However, in return for the increased cost, it was demonstrated that increased robustness followed, along with faster forward passes in the trained results. Our methods are orthogonal to the Jacobian stabilization process. In \Cref{sec:experiments}, we provide empirical evidence on composing our models with Jacobian Stabilization to achieve even more robust results.

\section{Methods}
\label{sec:methods}

\begin{table*}[t]
    \centering
    \adjustbox{max width=0.9\textwidth}{
        \centering
        \begin{tabular}{lcccccc}
            \toprule
            \thead{Model} & \thead{Jacobian Reg.} & \thead{\# of Params} & \thead{Test Accuracy (\%)} & \thead{Testing NFE} & \thead{Training Time (min)} & \thead{Prediction Time (s / batch)}\\
            \midrule
            Vanilla DEQ & \crossmark & 138K & \sdval{97.926}{0.107} & \sdval{18.345}{0.732} & \sdval{5.197}{1.106} & \sdval{0.038}{0.009} \\
                        & \tickmark  &      & \sdval{98.123}{0.025} & \sdval{\hp{0}5.034}{0.059}  & \sdval{7.321}{0.454} & \sdval{0.011}{0.005}\\
            \addlinespace
            Skip DEQ & \crossmark & 151K & \sdval{97.759}{0.080} & \sdval{\hp{0}4.001}{0.001} & \sdval{1.711}{0.202} & \sdval{0.010}{0.001}\\
                     & \tickmark  &      & \sdval{97.749}{0.141} & \sdval{\hp{0}4.001}{0.000} & \sdval{6.019}{0.234} & \sdval{0.012}{0.001}\\
            \addlinespace
            Skip Reg. DEQ & \crossmark & 138K & \sdval{97.973}{0.134} & \sdval{\hp{0}4.001}{0.000} & \sdval{1.295}{0.222} & \sdval{0.010}{0.001}\\
                        & \tickmark  &      & \sdval{98.016}{0.049} & \sdval{\hp{0}4.001}{0.000} & \sdval{5.128}{0.241} & \sdval{0.012}{0.000}\\
            \bottomrule
        \end{tabular}
    }
    \caption{\textbf{MNIST Classification with Fully Connected Layers}: Skip Reg. Continuous DEQ without Jacobian Regularization takes \textit{$\mathit{4\times}$ less training time} and \textit{speeds up prediction time by $4\times$} compared to Continuous DEQ. Continuous DEQ with Jacobian Regularization has a similar prediction time but takes \textit{$\mathit{6\times}$ more training time} than Skip Reg. Continuous DEQ. Using Skip variants \textit{speeds up training by $\mathit{1.42\times - 4\times}$}.}
    \label{tab:mnist_dense_summary}
\end{table*}

\subsection{Continuous Deep Equilibrium Networks}
\label{sec:continuous_deqs}

\begin{figure}[H]
    \centering
    \includegraphics[width=0.9\linewidth]{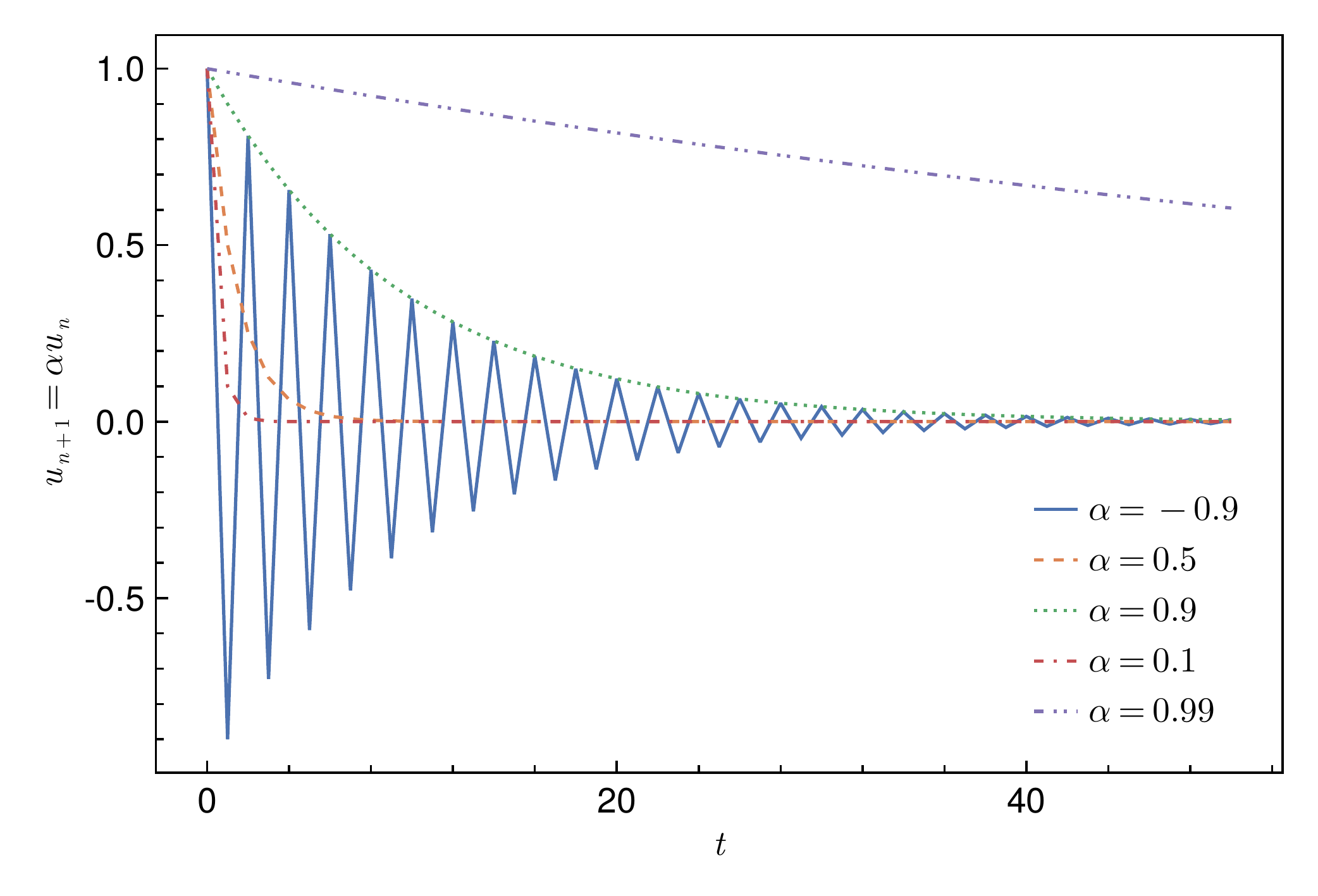}
    \caption{\textbf{Slow Convergence of Simple Linear Discrete Dynamical Systems}}
    \label{fig:linear_discrete_dynamical_system}
\end{figure}

Deep Equilibrium Models have traditionally been formulated as steady-state problems for a discrete dynamical system. However, discrete dynamical systems come with a variety of shortcomings. Consider the following linear discrete dynamical system (See \Cref{fig:linear_discrete_dynamical_system}):
\begin{equation}
    u_{n + 1} = \alpha \cdot u_n \qquad \texttt{ where } \|\alpha\| < 1 \texttt{ and } u_0 = 1
\end{equation}
This system converges to a steady state of $u_\infty = 0$. However, in many cases, this convergence can be relatively slow. If $\alpha = 0.9$, then after 10 steps, the value is $u_{10} = 0.35$ because a small amount only reduces each successive step. Thus convergence could only be accelerated by taking many steps together. Even further, if $\alpha = -0.9$, the value ping-pongs over the steady state $u_{1} = -0.9$, meaning that if we could take some fractional step $u_{\delta t}$ then it would be possible to approach the steady state much faster.
\citet{rico1992discrete, bulsari1995neural} describe several other shortcomings of using discrete steady-state dynamics over continuous steady-state dynamics. These issues combined motivate changing from a discrete description of the system (the fixed point or Broyden's method approach) to a continuous description of the system that allows adaptivity to change the stepping behavior and accelerate convergence.

To this end, we propose an alternate formulation for DEQs by modeling a continuous dynamical system (Continuous DEQ) where the forward pass is represented by an ODE which is solved from $t_0 = 0$ to $t_1 = \infty$:
\begin{equation}
    \frac{dz}{dt} = f_\theta(z, x) - z
\end{equation}
where $f_\theta$ is an explicit neural network. Continuous DEQs leverage fast adaptive ODE solvers, which terminate automatically once the solution is close to a steady state, i.e., $\frac{dz^*}{dt} = 0$, which then satisfies $f_\theta(z^*, x) = z^*$ and is thus the solution to the same implicit system as before.

The Continuous DEQ can be considered an infinite-time neural ODE in this form. However, almost paradoxically, the infinite time version is cheaper to train than the finite time version as its solution is the solution to the nonlinear system, meaning the same implicit differentiation formula of the original DEQ holds for the derivative. This means that no backpropagation through the steps is required for the Continuous DEQ, and only a linear system must be solved. In \Cref{sec:experiments}, we empirically demonstrate that Continuous DEQs outperform Neural ODEs in terms of training time while achieving similar accuracies.

\subsection{Skip Deep Equilibrium Networks}
\label{sec:skip_deqs}

\citet{bai_deep_2019, bai_multiscale_2020} set the initial condition $u_0 = 0$ while solving a DEQ. Assuming the existence of a steady state, the solvers will converge given enough iterations. However, each iteration is expensive, and a poor guess of the initial condition makes the convergence slower. To counteract these issues, we introduce an alternate architecture for DEQ (Skip~DEQ), where we use an explicit model $g_\phi$ to predict the initial condition for the steady-state problem $u_0 = g_\phi(x)$\footnote{We note that the concurrent work \citet{bai2021neural} introduced a similar formulation as a part of HyperDEQ}. We jointly optimize for $\left\{\theta, \phi\right\}$ by adding an auxiliary loss function:
\begin{equation}
    \mathcal{L}_{\texttt{skip}} = \lambda_{\texttt{skip}} \| f_\theta(z^*, x) - g_\phi(x) \|
\end{equation}
Intuitively, our explicit model $g_\phi$ better predicts a value closer to the steady-state (over the training iterations), and hence we need to perform fewer iterations during the forward pass. Given that its prediction is relatively free compared to the cost of the DEQ, this technique could decrease the cost of the DEQ by reducing the total number of iterations required. However, this prediction-correction approach still uses the result of the DEQ for its final predictions and thus should achieve robustness properties equal.

\subsubsection{Skip Regularized DEQ: Regularization Scheme without Extra Parameters}
\label{subsubsec:skip_reg_deq}

One of the primary benefits of DEQs is the low memory footprint of these models (See \Cref{sec:background}). Introducing an explicit model $g_\phi$ increases the memory requirements for training. To alleviate this problem, we propose a regularization term to minimize the L1 distance between the first prediction of $f_\theta$ and the steady-state solution:
\begin{align}
    \mathcal{L}_{\texttt{skip}} = \lambda_{\texttt{skip}} \| f_\theta(z^*, x) - f_\theta(0, x) \|
\end{align}
This technique follows the same principle as the Skip DEQ where the DEQ's internal neural network is now treated as the prediction model. We hypothesize that this introduces an inductive bias in the model to learn simpler training dynamics.

\begin{table*}[t]
    \centering
    \adjustbox{max width=0.9\textwidth}{
        \centering
        \begin{tabular}{lcccccc}
            \toprule
            \thead{Model} & \thead{Continuous} & \thead{\# of Params} & \thead{Test Accuracy (\%)} & \thead{Training Time\\ (s / batch)} & \thead{Backward Pass\\ (s / batch)} & \thead{Prediction Time\\ (s / batch)}\\
            \midrule
            Vanilla DEQ & \crossmark & 163546 & \sdval{81.233}{0.097} & \sdval{0.651}{0.009} & \sdval{0.075}{0.001} & \sdval{0.282}{0.005}\\
                        & \tickmark  &        & \sdval{80.807}{0.631} & \sdval{0.753}{0.017} & \sdval{0.261}{0.010} & \sdval{0.136}{0.010}\\
            \addlinespace
            Skip DEQ & \crossmark & 200122 & \sdval{82.013}{0.306} & \sdval{0.717}{0.022} & \sdval{0.115}{0.004} & \sdval{0.274}{0.005}\\
                     & \tickmark  &        & \sdval{80.807}{0.230} & \sdval{0.806}{0.010} & \sdval{0.293}{0.004} & \sdval{0.154}{0.002}\\
            \addlinespace
            Skip Reg. DEQ & \crossmark & 163546 & \sdval{81.170}{0.356} & \sdval{0.709}{0.005} & \sdval{0.114}{0.002} & \sdval{0.283}{0.007}\\
                        & \tickmark  &        & \sdval{82.513}{0.177} & \sdval{0.679}{0.015} & \sdval{0.143}{0.017} & \sdval{0.154}{0.003}\\
            \addlinespace
            Neural ODE  & \tickmark  & 163546 & \sdval{83.543}{0.393} & \sdval{1.608}{0.026} & \sdval{1.240}{0.021} & \sdval{0.207}{0.006}\\
            \bottomrule
        \end{tabular}
    }
    \caption{\textbf{CIFAR10 Classification with Small Neural Network}: Skip Reg. Continuous DEQ achieves the \textit{highest test accuracy among DEQs}. Continuous DEQs are faster than Neural ODEs during training by a factor of $\mathit{2\times - 2.36\times}$, with a speedup of $\mathit{4.2\times - 8.67\times}$ in the backward pass. We also observe a prediction speed-up for Continuous DEQs of $\mathit{1.77\times - 2.07\times}$ against Discrete DEQs and $\mathit{1.34\times - 1.52\times}$ against Neural ODE.}
    \label{tab:cifar10_tiny_summary}
\end{table*}

\begin{figure*}[t]
    \centering
    \includegraphics[width=0.9\linewidth]{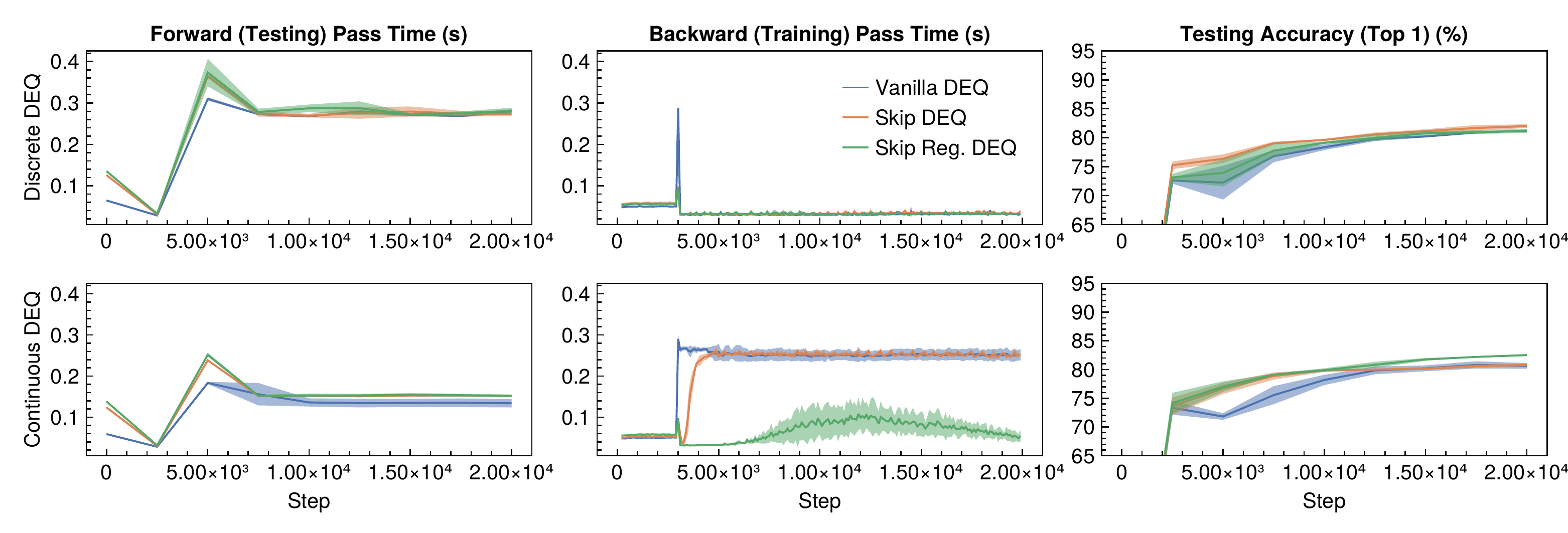}
    \vspace{-1em}
    \caption{\textbf{CIFAR10 Classification with Small Neural Network}}
    \label{fig:cifar10_tiny}
\end{figure*}

\begin{table*}[t]
    \centering
    \adjustbox{max width=0.9\textwidth}{
        \centering
        \begin{tabular}{lcccccc}
            \toprule
            \thead{Model} & \thead{Continuous} & \thead{\# of Params} & \thead{Test Accuracy (\%)} & \thead{Training Time\\ (s / batch)} & \thead{Backward Pass\\ (s / batch)} & \thead{Prediction Time\\ (s / batch)}\\
            \midrule
            Vanilla DEQ & \crossmark & 10.63M & \sdval{88.913}{0.287} & \sdval{0.625}{0.165} & \sdval{0.111}{0.021} & \sdval{0.414}{0.222}\\
                        & \tickmark  &        & \sdval{89.367}{0.832} & \sdval{1.284}{0.011} & \sdval{0.739}{0.003} & \sdval{0.606}{0.010}\\
            \addlinespace
            Skip DEQ    & \crossmark & 11.19M & \sdval{88.783}{0.178} & \sdval{0.588}{0.042} & \sdval{0.112}{0.006} & \sdval{0.314}{0.017}\\
                        & \tickmark  &        & \sdval{89.600}{0.947} & \sdval{0.697}{0.012} & \sdval{0.150}{0.013} & \sdval{0.625}{0.004}\\
            \addlinespace
            Skip Reg. DEQ & \crossmark & 10.63M & \sdval{88.773}{0.115} & \sdval{0.613}{0.048} & \sdval{0.109}{0.008} & \sdval{0.268}{0.031}\\
                        & \tickmark  &        & \sdval{90.107}{0.837} & \sdval{0.660}{0.019} & \sdval{0.125}{0.003} & \sdval{0.634}{0.019}\\
            \addlinespace
            Neural ODE  & \tickmark & 10.63M  & \sdval{89.047}{0.116} & \sdval{5.267}{0.078} & \sdval{4.569}{0.077} & \sdval{0.573}{0.010}\\
            \bottomrule
        \end{tabular}
    }
    \caption{\textbf{CIFAR10 Classification with Large Neural Network}: Skip Reg. Continuous DEQ achieves the \textit{highest test accuracy}. Continuous DEQs are faster than Neural ODEs during training by a factor of \timeschange{4.1}{7.98}, with a speedup of \timeschange{6.18}{36.552} in the backward pass. However, we observe a prediction slowdown for Continuous DEQs of \timeschange{1.4}{2.36} against Discrete DEQs and \timeschange{0.90}{0.95} against Neural ODE.}
    \label{tab:cifar10_large_summary}
\end{table*}

\begin{figure*}[t]
    \centering
    \includegraphics[width=0.9\linewidth]{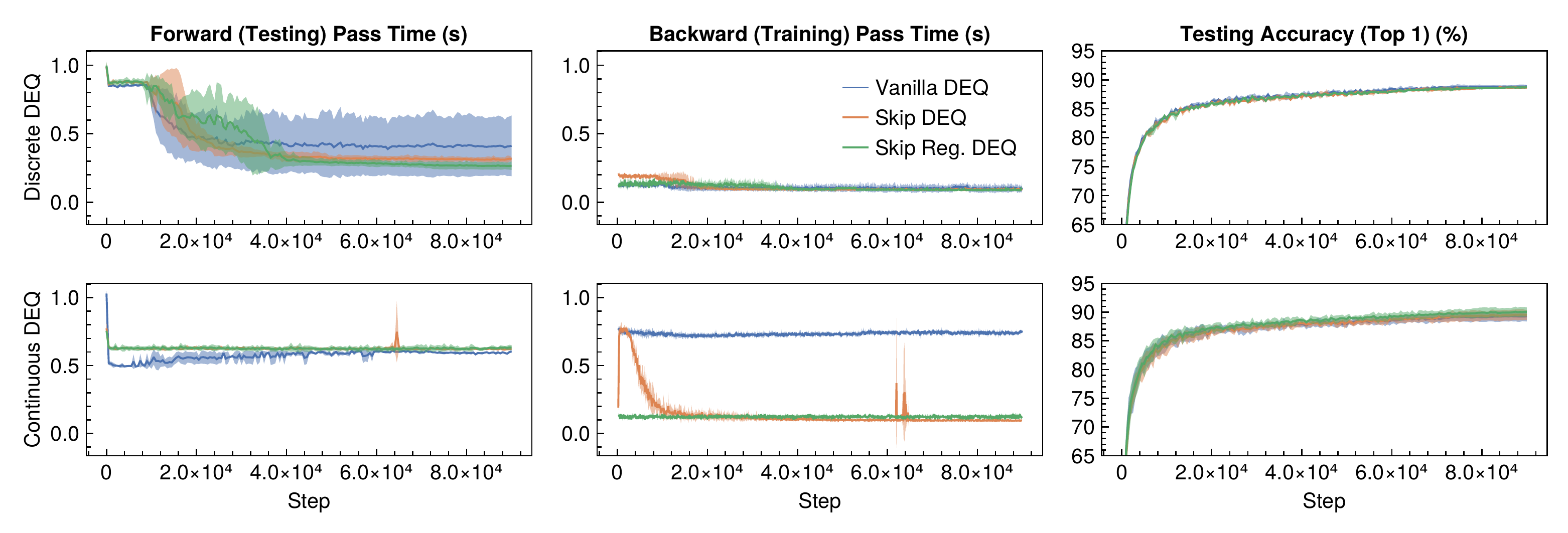}
    \vspace{-1em}
    \caption{\textbf{CIFAR10 Classification with Large Neural Network}}
    \label{fig:cifar10_large}
\end{figure*}

\section{Experiments}
\label{sec:experiments}

In this section, we consider the effectiveness of our proposed methods -- Continuous DEQs and Skip DEQs -- on the training and prediction timings. We consider the following baselines:
\begin{enumerate}
    \item Discrete DEQs with L-Broyden Solver.
    \item Jacobian Regularization of DEQs.\footnote{We note that due to limitations of our Automatic Differentiation system, we cannot perform Jacobian Regularization for Convolutional Models. However, our preliminary analysis suggests that the Skip DEQ and Continuous DEQ approaches are fully composable with Jacobian Regularization and provide better performance compared to using only Jacobian Regularization (See \Cref{tab:mnist_dense_summary}).}
    \item Multi-Scale Neural ODEs with Input Injection: A modified Continuous Multiscale DEQ without the steady state convergence constaint.
\end{enumerate}
Our primary metrics are classification accuracy, the number of function evaluations (NFEs), total training time, time for the backward pass, and prediction time per batch. We showcase the performance of our methods on -- MNIST~\citep{lecun1998gradient}, CIFAR-10~\citep{krizhevsky2009learning}, SVHN~\citep{netzer2011reading}, \& ImageNet~\citep{deng2009imagenet}. We use perform our experiments in Julia~\citep{Julia-2017} using Lux.jl~\citep{pal2022lux} and DifferentialEquations.jl~\citep{DifferentialEquations.jl-2017, rackauckas2018comparison, rackauckas2020universal}.

\subsection{MNIST Image Classification}
\label{subsec:mnist_image_classification}

\textbf{Training Details:} Following \citet{kelly2020learning}, our Fully Connected Model consists of 3 layers -- a downsampling layer $\mathbb{R}^{784} \mapsto \mathbb{R}^{128}$, continuous DEQ layer $f_\theta: \mathbb{R}^{128} \mapsto \mathbb{R}^{128}$, and a linear classifier $\mathbb{R}^{128} \mapsto \mathbb{R}^{10}$.

For regularization, we use $\lambda_{\texttt{skip}} = 0.01$ and train the models for $25$ epochs with a batch size of $32$. We use Tsit5~\citep{tsitouras2011runge} with a relative tolerance for convergence of $0.005$. For optimization, we use Adam~\citep{kingma2014adam} with a constant learning rate of $0.001$.

\textbf{Baselines:} We use continuous DEQ and continuous DEQ with Jacobian Stabilization as our baselines. We additionally compose Skip DEQs with Jacobian Stabilization in our benchmarks. For all experiments, we keep $\lambda_{\texttt{jac}} = 1.0$.

\textbf{Results:} We summarize our results in \Cref{tab:mnist_dense_summary}. Without Jacobian Stabilization, Skip Reg. Continuous DEQ has the highest testing accuracy of $\mathit{97.973\%}$ and has the \textit{lowest training and prediction timings overall}. Using Jacobian Regularization, DEQ outperforms Skip~DEQ models by $\mathit{< 0.4\%}$, however, jacobian regularization increases training time by $\mathit{1.4 - 4}\times$. Skip~DEQ models can obtain the lowest prediction time per batch of $\mathit{\sim0.01s}$.

\subsection{CIFAR10 Image Classification}
\label{subsec:cifar10_image_classification}

For all the baselines in this section, Vanilla DEQ is trained with the same training hyperparameters as the corresponding Skip DEQs (taken from \citet{bai_multiscale_2020}). Multiscale Neural ODE with Input Injection is trained with the same hyperparameters as the corresponding Continuous DEQs.

\subsubsection{Architecture with ~200K parameters}

\begin{table*}[t]
    \centering
    \adjustbox{max width=0.9\textwidth}{
        \centering
        \begin{tabular}{lcccccc}
            \toprule
            \thead{Model} & \thead{Continuous} & \thead{\# of Params} & \thead{Test Accuracy\\ (Top 5) (\%)} & \thead{Training Time\\ (s / batch)} & \thead{Backward Pass\\ (s / batch)} & \thead{Prediction Time\\ (s / batch)}\\
            \midrule
            Vanilla DEQ & \crossmark & 17.91M & \sdval{81.809}{0.115} & \sdval{2.057}{0.138} & \sdval{0.195}{0.007} & \sdval{1.963}{0.189}\\
                        & \tickmark  &        & \sdval{81.329}{0.516} & \sdval{3.131}{0.027} & \sdval{1.873}{0.015} & \sdval{1.506}{0.027}\\
            \addlinespace
            Skip DEQ    & \crossmark & 18.47M & \sdval{81.717}{0.452} & \sdval{1.956}{0.012} & \sdval{0.194}{0.001} & \sdval{1.843}{0.025}\\
                        & \tickmark  &        & \sdval{81.334}{0.322} & \sdval{2.016}{0.129} & \sdval{0.845}{0.127} & \sdval{1.575}{0.053}\\
            \addlinespace
            Skip Reg. DEQ & \crossmark & 17.91M & \sdval{81.611}{0.369} & \sdval{1.996}{0.035} & \sdval{0.539}{0.023} & \sdval{1.752}{0.093}\\
                        & \tickmark  &        & \sdval{81.813}{0.350} & \sdval{1.607}{0.044} & \sdval{0.444}{0.026} & \sdval{1.560}{0.021}\\
            \bottomrule
        \end{tabular}
    }
    \caption{\textbf{ImageNet Classification}: All the variants attain comparable evaluation accuracies. Skip (Reg.) accelerates the training of Continuous DEQ by $\mathit{1.57\times - 1.96\times}$, with a reduction of $\mathit{2.2\times - 4.2\times}$ in the backward pass timings. However, we observe a marginal increase of $\mathit{4\%}$ in prediction timings for Skip (Reg.) Continuous DEQ compared against Continuous DEQ. For Discrete DEQs, Skip (Reg.) variants reduce the prediction timings by $\mathit{6.5\% - 12\%}$.}
    \label{tab:imagenet_summary}
\end{table*}

\begin{figure*}[t]
    \centering
    \includegraphics[width=0.9\linewidth]{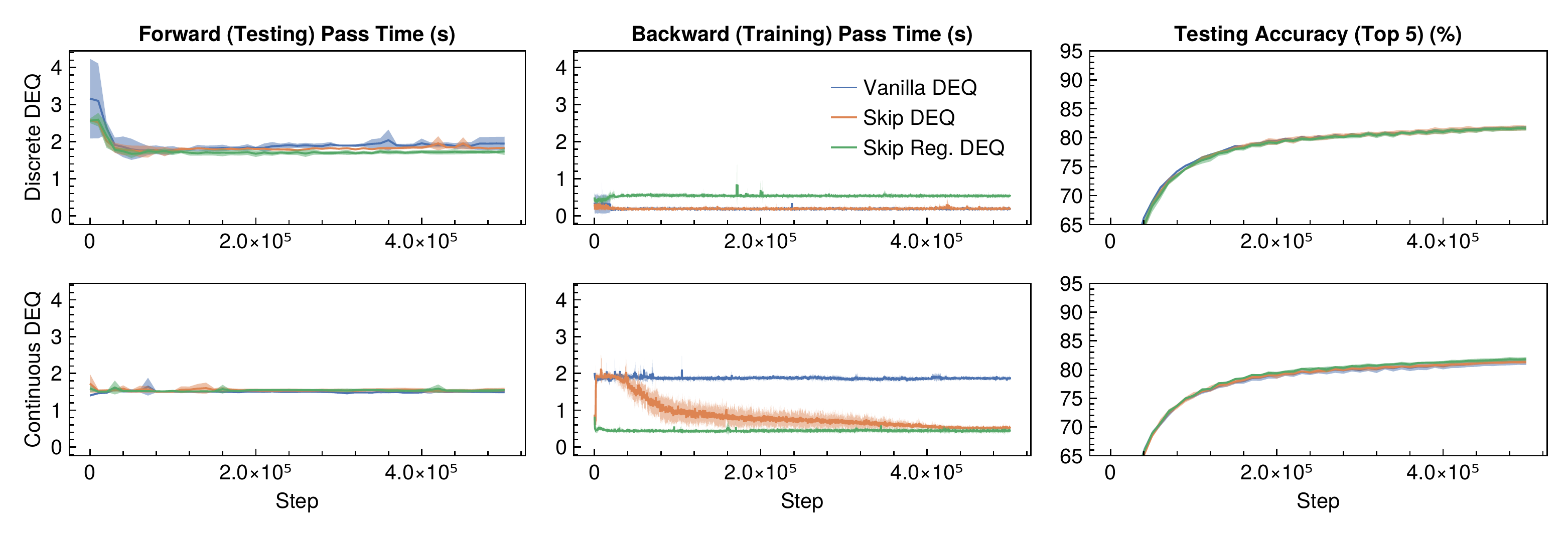}
    \vspace{-1em}
    \caption{\textbf{ImageNet Classification}}
    \vspace{-1em}
    \label{fig:imagenet_small}
\end{figure*}

\textbf{Training Details:} Our Multiscale DEQ architecture is the same as MDEQ-small architecture used in \citet{bai_multiscale_2020}. For the explicit network in Skip~DEQ, we use the residual block and downsampling blocks from \citet{bai_multiscale_2020} which account for the additional 58K trainable parameters.

We use a fixed regularization weight of $\lambda_{\texttt{skip}} = 0.01$ and the models are trained for 20000 steps. We use a batch size of $128$. For continuous models, we use VCAB3~\citep{wanner1996solving} with a relative tolerance for convergence of $0.05$. We use AdamW~\citep{loshchilov2017decoupled} optimizer with a cosine scheduling on the learning rate -- starting from $10^{-3}$ and terminating at $10^{-6}$ -- and a weight decay of $2.5 \times 10^{-6}$.

\textbf{Results:} We summarize our results in \Cref{tab:cifar10_tiny_summary} and \Cref{fig:cifar10_tiny}. Continuous DEQs are faster than Neural ODEs during training by a factor of $\mathit{2\times - 2.36\times}$, with a speedup of $\mathit{4.2\times - 8.67\times}$ in the backward pass.

\subsubsection{Architecture with 11M parameters}

\textbf{Training Details:} Our Multiscale DEQ architecture is the same as MDEQ-large architecture used in \citet{bai_multiscale_2020}. For the explicit network in Skip~DEQ, we use the residual block and downsampling blocks from \citet{bai_multiscale_2020} which account for the additional 58K trainable parameters.

We use a fixed regularization weight of $\lambda_{\texttt{skip}} = 0.01$ and the models are trained for 90000 steps. We use a batch size of $128$. For continuous models, we use VCAB3~\citep{wanner1996solving} with a relative tolerance for convergence of $0.05$. We use Adam~\citep{kingma2014adam} optimizer with a cosine scheduling on the learning rate -- starting from $10^{-3}$ and terminating at $10^{-6}$.

\textbf{Results:} We summarize our results in \Cref{tab:cifar10_large_summary} and \Cref{fig:cifar10_large}. Continuous DEQs are faster than Neural ODEs during training by a factor of \timeschange{4.1}{7.98}, with a speedup of \timeschange{6.18}{36.552} in the backward pass.

\subsection{ImageNet Image Classification}

\textbf{Training Details:} Our Multiscale DEQ architecture is the same as MDEQ-small architecture used in \citet{bai_multiscale_2020}. For the explicit network in Skip~DEQ, we use the residual block and downsampling blocks from \citet{bai_multiscale_2020} which account for the additional 58K trainable parameters.

We use a fixed regularization weight of $\lambda_{\texttt{skip}} = 0.01$, and the models are trained for 500000 steps. We use a batch size of $64$. For continuous models, we use VCAB3~\citep{wanner1996solving} with a relative tolerance for convergence of $0.05$. We use SGD with a momentum of $0.9$ and weight decay of $10^{-6}$. We use a step LR scheduling reducing the learning rate from $0.05$ by a multiplicative factor of $0.1$ at steps $100000$, $150000$, and $250000$.

\textbf{Baselines:} Vanilla DEQ is trained with the same training hyperparameters as the corresponding Skip DEQs (taken from \citep{bai_multiscale_2020})\footnote{When training MultiScale Neural ODE with the same configuration as Continuous DEQ, we observed a $\mathit{8\times}$ slower backward pass which made the training of the baseline infeasible.}.

\textbf{Results:} We summarize our results in \Cref{tab:imagenet_summary} and \Cref{fig:imagenet_small}. Skip (Reg.) variants accelerate the training of Continuous DEQ by $\mathit{1.57\times - 1.96\times}$, with a reduction of $\mathit{2.2\times - 4.2\times}$ in the backward pass timings.

\section{Related Works}
\label{sec:related_works}

\subsection{Implicit Models}

Implicit Models have obtained competitive results in image processing~\citep{bai_multiscale_2020}, generative modeling~\citep{grathwohl2018ffjord}, time-series prediction~\citep{rubanova2019latent}, etc, at a fraction of memory requirements for explicit models. Additionally, \citet{kawaguchi2021theory} show that for a certain class of DEQs convergence to global optima is guaranteed at a linear rate. However, the slow training and prediction timings~\citep{dupont2019augmented, kelly2020learning, finlay2020train, ghosh2020steer, pmlr-v139-pal21a, bai2021stabilizing} often overshadow these benefits.

\vspace{-1em}

\subsection{Accelerating Neural ODEs}

\citet{finlay2020train, kelly2020learning} used higher-order regularization terms to constrain the space of learnable dynamics for Neural ODEs. Despite speeding up predictions, these models often increase the training time by 7x~\citep{pmlr-v139-pal21a}. Alternatively, \citet{ghosh2020steer} randomized the endpoint of Neural ODEs to incentivize simpler dynamics. \citet{pmlr-v139-pal21a} used internal solver heuristics -- local error and stiffness estimates -- to control the learned dynamics in a way that decreased both prediction and training time. \citet{xia2021heavy} rewrite Neural ODEs as heavy ball ODEs to accelerate both forward and backward passes. \citet{djeumou2022taylorlagrange} replace ODE solvers in the forward with a Taylor-Lagrange expansion and report significantly better training and prediction times.

Regularized Neural ODEs can not be directly extended to discrete DEQs~\citep{bai_deep_2019, bai_multiscale_2020}. Our continuous formulation introduces the potential to extend \citet{xia2021heavy, djeumou2022taylorlagrange} to DEQs. However, these methods benefit from the structure in the backward pass, which does not apply to DEQs. Additionally, relying on discrete sensitivity analysis~\citep{pmlr-v139-pal21a} nullifies the benefit of a cost-effective backward pass. 

\subsection{Accelerating DEQs}

\citet{bai2021stabilizing} uses second-order derivatives to regularize the Jacobian, stabilizing the training and prediction timings of DEQs. \citet{fung2022jfb} proposes a Jacobian-Free Backpropagation Model, which accelerates solving the Linear Equation in the backward pass. Our work complements these models and can be freely composed with them. We have shown that a poor initial condition harms convergence, and a better estimate for the same leads to faster training and prediction. We hypothesize that combining these methods would lead to more stable and faster convergence, demonstrating this possibility with the Jacobian regularization Skip DEQ.

\section{Discussion}
\label{sec:discussion}

We have empirically shown the effectiveness of Continuous~DEQs as a faster alternative for Neural ODEs. Consistent with the ablation studies in \citet{bai2021neural}, we see that Skip DEQ in itself doesn't significantly improve the prediction or training timings for Discrete DEQs. Skip Reg. DEQ does, however, speeds up the inference for larger Discrete DEQs. However, combining Skip DEQ and Skip Reg. DEQ with Continuous DEQs, enable a speedup in backward pass by over $\mathit{2.8} - \mathit{5.9 \times}$. We hypothesize that this improvement is due to reduction in the condition number, which results in faster convergence of GMRES in the backward pass, however, acertaining this would require furthur investigation. We have demonstrated that our improvements to DEQs and Neural ODEs enable the drop-in replacement of Skip Continuous DEQs in any classical deep learning problem where continuous implicit models were previously employed.

\subsection{Limitations}
\label{sec:limitations}

We observe the following limitations for our proposed methods:
\begin{itemize}
    \item Reformulating a Neural ODE as a Continuous DEQ is valid, when the actual dynamics of the system doesn't matter. This holds true for all applications of Neural ODEs to classical Deep Learning problems.

    \item Continuous DEQs are slower than their Discrete counterparts for larger models (without any significant improvement to accuracy), hence the authors recommend their usage only for cases where a continuous model is truly needed.
\end{itemize}

\section{Acknowledgement}
\label{sec:acknowledgement}

The authors acknowledge the MIT SuperCloud and Lincoln Laboratory Supercomputing Center for providing HPC resources that have contributed to the research results reported within this paper. This material is based upon work supported by the National Science Foundation under grant no. OAC-1835443, grant no. SII-2029670, grant no. ECCS-2029670, grant no. OAC-2103804, and grant no. PHY-2021825. We also gratefully acknowledge the U.S. Agency for International Development through Penn State for grant no. S002283-USAID. The information, data, or work presented herein was funded in part by the Advanced Research Projects Agency-Energy (ARPA-E), U.S. Department of Energy, under Award Number DE-AR0001211 and DE-AR0001222. We also gratefully acknowledge the U.S. Agency for International Development through Penn State for grant no. S002283-USAID. The views and opinions of authors expressed herein do not necessarily state or reflect those of the United States Government or any agency thereof. This material was supported by The Research Council of Norway and Equinor ASA through Research Council project ``308817 - Digital wells for optimal production and drainage''. Research was sponsored by the United States Air Force Research Laboratory and the United States Air Force Artificial Intelligence Accelerator and was accomplished under Cooperative Agreement Number FA8750-19-2-1000. The views and conclusions contained in this document are those of the authors and should not be interpreted as representing the official policies, either expressed or implied, of the United States Air Force or the U.S. Government. The U.S. Government is authorized to reproduce and distribute reprints for Government purposes notwithstanding any copyright notation herein.

\bibliography{ref}
\bibliographystyle{icml2023}

\end{document}